\documentclass[10pt,twocolumn,letterpaper]{article}
\usepackage{wacv}
\usepackage{times}
\usepackage{epsfig}
\usepackage{graphicx}
\usepackage{amsmath}
\usepackage{amssymb}
\usepackage{enumitem}
\usepackage{subcaption}
\usepackage[table]{xcolor}
\usepackage{multirow}

\newcommand{\by}{\mathbf{y}}

\def\wacvPaperID{106} 

\wacvfinalcopy 

\ifwacvfinal
\usepackage[breaklinks=true,bookmarks=false]{hyperref}
\else
\usepackage[pagebackref=true,breaklinks=true,colorlinks,bookmarks=false]{hyperref}
\fi

\begin{document}

\title{AT-DDPM: Restoring Faces degraded by Atmospheric Turbulence using Denoising Diffusion Probabilistic Models}
\author{Nithin Gopalakrishnan Nair\\
Johns Hopkins University\\
{\tt\small ngopala2@jhu.edu}
\and
Kangfu Mei\\
Johns Hopkins University\\
{\tt\small kmei1@jhu.edu}
\and
Vishal M. Patel\\
Johns Hopkins University\\
{\tt\small vpatel36@jhu.edu}
}

\twocolumn[{%
\renewcommand\twocolumn[1][]{#1}%
\maketitle
\vspace{1.5cm}
\vspace{-3\baselineskip}
\vspace{-3\baselineskip}
\begin{center}
\centering
\setlength{\tabcolsep}{0.5pt}
\captionsetup{type=figure}
{\small
\renewcommand{\arraystretch}{0.5} 
\begin{tabular}{c c c c c c c c c}
  \includegraphics[width=0.12\linewidth]{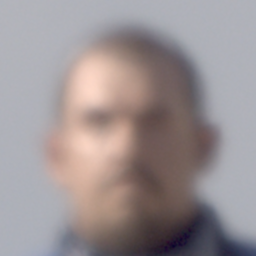}
    \includegraphics[width=0.12\linewidth]{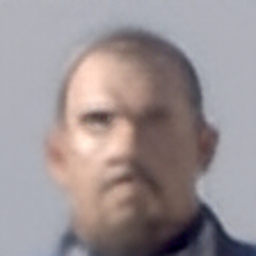}
    \includegraphics[width=0.12\linewidth]{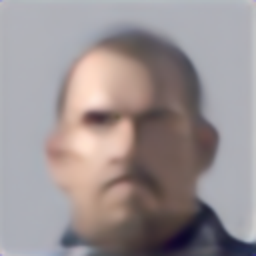}
    \includegraphics[width=0.12\linewidth]{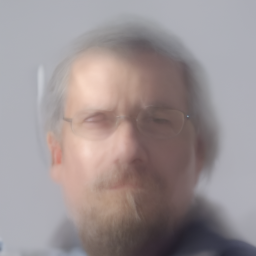}
    \includegraphics[width=0.12\linewidth]{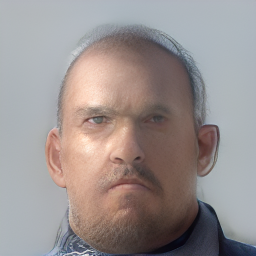}
    \includegraphics[width=0.12\linewidth]{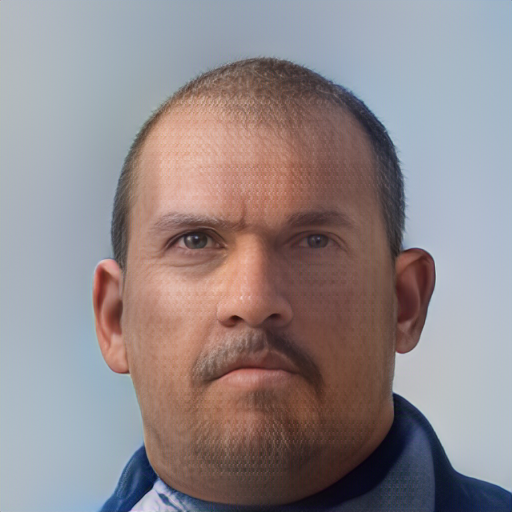}
  \includegraphics[width=0.12\linewidth]{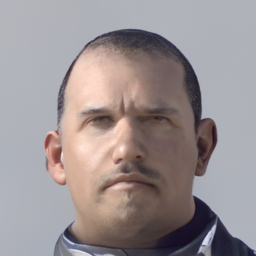}
    \includegraphics[width=0.12\linewidth]{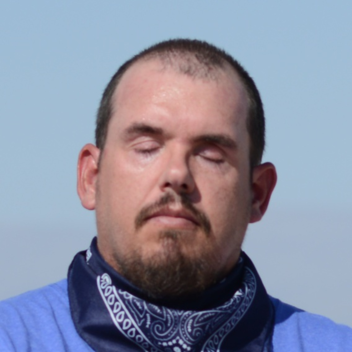}
    \tabularnewline
   \includegraphics[width=0.12\linewidth]{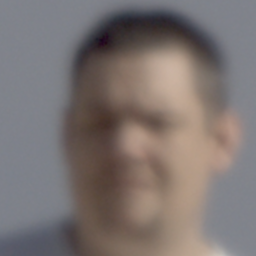}
    \includegraphics[width=0.12\linewidth]{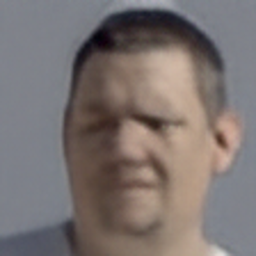}
    \includegraphics[width=0.12\linewidth]{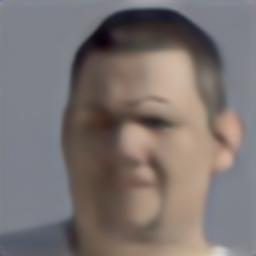}
    \includegraphics[width=0.12\linewidth]{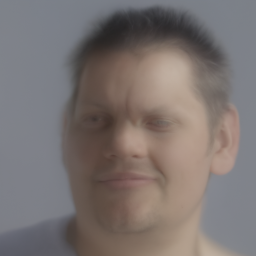}
    \includegraphics[width=0.12\linewidth]{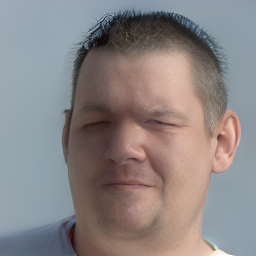}
    \includegraphics[width=0.12\linewidth]{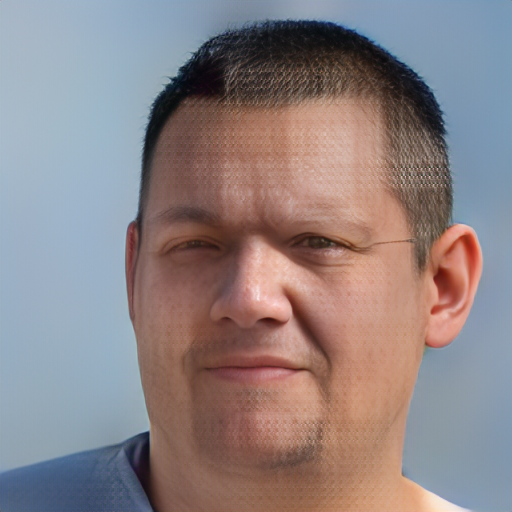}
  \includegraphics[width=0.12\linewidth]{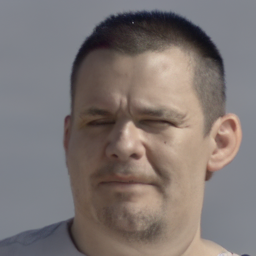}
    \includegraphics[width=0.12\linewidth]{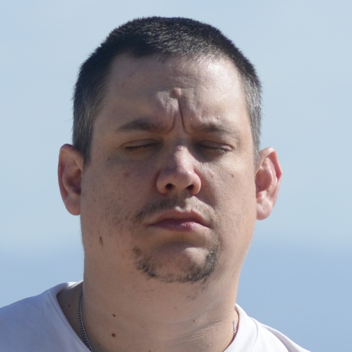}
    \tabularnewline
   \includegraphics[width=0.12\linewidth]{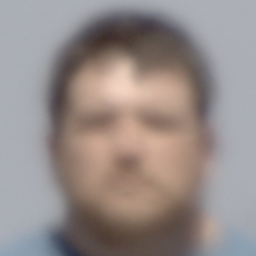}
    \includegraphics[width=0.12\linewidth]{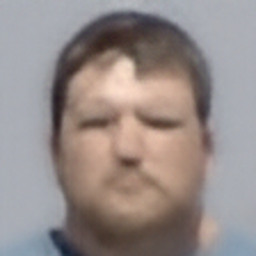}
    \includegraphics[width=0.12\linewidth]{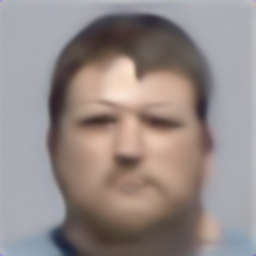}
    \includegraphics[width=0.12\linewidth]{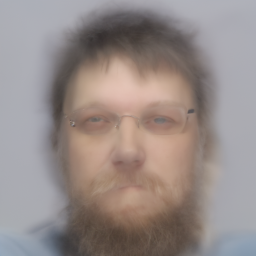}
    \includegraphics[width=0.12\linewidth]{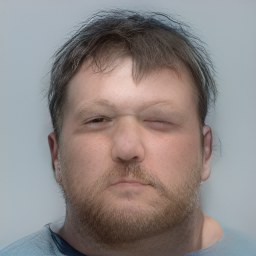}
    \includegraphics[width=0.12\linewidth]{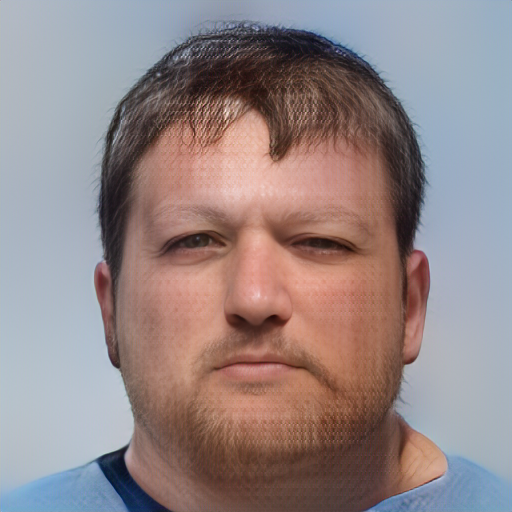}
  \includegraphics[width=0.12\linewidth]{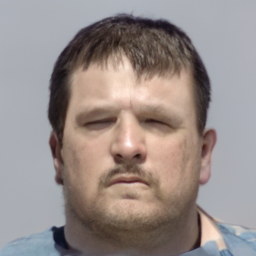}
    \includegraphics[width=0.12\linewidth]{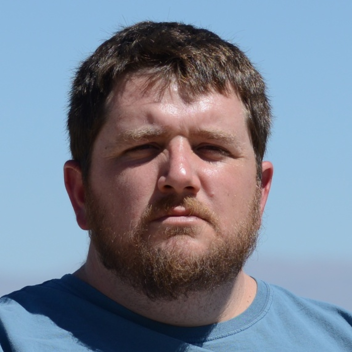}
    \tabularnewline
{\hspace{-5pt}Distorted\hskip20pt ATFaceGAN\cite{lau2020atfacegan}\hskip15pt ATNet\cite{yasarla2021learning}\hskip15pt ILVR\cite{choi2021ilvr}\hskip25pt GFPGAN\cite{wang2021towards}\hskip15pt LTTGAN~\cite{mei2021ltt}\hskip20pt OURS \hskip30pt Gallery}\\
\end{tabular}}
\vspace{-0.5\baselineskip}
\captionof{figure}{Sample qualitative comparisons with single image atmospheric turbulence mitigation methods as well as generative restoration models on the LRFID\cite{miller2019data} dataset.}
\label{fig:facelrfid}
\vspace{-2mm}
\end{center}%
}]
\thispagestyle{empty}

\begin{abstract}
\vspace{-5mm}
Although many long-range imaging systems are designed to support extended vision applications, a natural obstacle to their operation is degradation due to atmospheric turbulence.  Atmospheric turbulence causes significant degradation to image quality by introducing blur and geometric distortion. In recent years, various deep learning-based single image atmospheric turbulence mitigation methods, including CNN-based and GAN inversion-based, have been proposed in the literature which attempt to remove the distortion in the image. However, some of these methods are difficult to train and often fail to reconstruct  facial features and produce unrealistic results especially in the case of high turbulence. 
Denoising Diffusion Probabilistic Models (DDPMs) have recently gained some traction because of their stable training process and their ability to generate high quality images. In this paper, we propose the first DDPM-based solution for the problem of atmospheric turbulence mitigation. We also propose a fast sampling technique for reducing the inference times for conditional DDPMs. Extensive experiments are conducted on synthetic and real-world data to show the significance of our model. To facilitate further research, all codes and pretrained models are publically available at http://github.com/Nithin-GK/AT-DDPM
\end{abstract}
\vspace{-7mm}
\section{Introduction}
\vspace{-2mm}
  Autonomous vehicles and surveillance systems utilize long-range imaging systems to capture faraway scenes. Images captured by such systems are prone to degradation due to variations of the refractive index of air. Atmospheric turbulence is a ubiquitous phenomenon that causes spatially and temporally random variations in the air's refractive index  \cite{roggemann2018imaging,hufnagel1964modulation}. This introduces random geometric distortions and blur to the images captured and significantly degrades the performance of computer vision tasks like object detection \cite{oreifej2012simultaneous}, tracking and recognition \cite{chen2014detecting} applied on these images. Modeling atmospheric turbulence degradation is an extremely complex task. Various methods have been proposed in the literature that attempt to model deformations caused due to turbulence.  Under the assumption that the scene and the imaging sensor are both static and degradations caused are due to atmospheric turbulence alone, the degradation due to atmospheric turbulence is modelled as 
  \setlength{\belowdisplayskip}{0pt} \setlength{\belowdisplayshortskip}{0pt}
\setlength{\abovedisplayskip}{0pt} \setlength{\abovedisplayshortskip}{0pt}
 \begin{equation}
      T_k = D_k(H_k(I)))+n_k, \;\;\;k=1,2,...,N,
      \label{eq: turb}
 \end{equation}
 where $T_k$ is the degraded image at the $k^{th}$ time instant, $I$ is the clean latent image, $D_k$ is the deformation operator which is assumed to deform randomly, $n_k$ is additive noise, and $H_k$ is air turbulence-caused  blurring  operator \cite{lau2021atfacegan,9320575,zhu2012removing,belen2001turbulence,hirsch2010efficient} . 
 
Mitigating the effect of atmospheric turbulence and restoring the latent clean image is extremely challenging due to its ill-posed nature that arises due to both blur and geometric distortions.  Restoring images affected by atmospheric turbulence is an extensively researched topic in the literature. To mitigate these effects, there exists two classical solutions -- adaptive optics  \cite{pearson1976atmospheric,tyson2015principles,roggemann2018imaging} and image processing \cite{furhad2016restoring,zhu2012removing,anantrasirichai2013atmospheric,vorontsov2001anisoplanatic,aubailly2009automated,meinhardt2014implementation,lau2019restoration}. In this paper we focus on the latter and propose an image processing-based strategy for atmospheric turbulence mitigation. 

In recent years, various deep learning-based methods for mitigating atmospheric turbulence have been proposed in the literature \cite{lau2021atfacegan,lau2019restoration,yasarla2020learning}. \cite{yasarla2020learning} proposes a deep network utilizing blur and degradation priors to restore the clean image.  \cite{lau2021atfacegan} uses two parallel paths to remove blur and degradation and finally concatenates them to remove the effect of turbulence. These methods tackle atmospheric turbulence degradation on facial images but  focus on removing the distortion and does not have enough generative power to generate a realistic face for strong turbulence distortions. Moreover these methods fail to generalize well in case of real world images. Generative prior-based face reconstruction has gained much attention in recent years due to their ability in modelling the distribution of clean facial images \cite{wang2021towards,mei2021ltt}. This is usually  performed by utilizing Generative Adversarial Networks (GANs) for modelling the distribution of clean faces and then utilizing these GANs as a pretrained prior to generate realistic facial images while performing restoration. But training of GANs is unstable and may also fail due to mode collapse. In this paper, we propose a generative modeling-based solution using Denoising Diffusion Probabilistic Models (DDPMs)\cite{sohl2015deep,ho2020denoising}. DDPM-based methods have already achieved state-of-the-art results in image generation tasks \cite{dhariwal2021diffusion}. When compared to GANs, their training process is also much more stable\cite{ho2020denoising,dhariwal2021diffusion}. Motivated by their ability to perform exact sampling and generating high quality images, we use them as generative priors to restore single facial images degraded by atmospheric turbulence.
We first train a DDPM model to learn the  distribution of facial images by training on a large dataset\cite{karras2019style}. We then adapt this model to learn the transformation for the simpler restoration problem of image super resolution by using knowledge distillation. With the super-resolution model in hand, we adapt these weights again for the transformation from turbulent images to clean faces by performing knowledge distillation with the super-resolution model. The could be thought of as a continual learning based strategy\cite{zenke2017continual}. Where, the final model has access to information about the manifold of realistic faces, hence being able to produce realistic face outputs even for strong distortions. Restoration of images degraded by atmospheric turbulence is an ill-posed problem.  Hence, multiple possibilities of clean faces exist while reconstructing using a DDPM model that has access to the distribution of clean facial images. 
Moreover, the inference process of DDPMs is quite slow and time consuming.
To this end we enforce a constraint to ensure that the facial features present in the restored image are closer to the ones present in the distorted image during inference by starting from noise turbulence-distorted image rather than pure Gaussian noise. 

This paper makes the following contributions:
\begin{itemize}
    \item We propose the first generative prior-based image restoration method using DDPM for the task of atmospheric turbulence mitigation.
    \item We propose a progressive training(PT)-based learning framework for better reconstruction and introduce a new sampling procedure for efficient inference.
    \item We qualitatively and quantitatively evaluate our method on both synthetic and real-world datasets and show that it performs better than the existing state-of-the-art single image-based atmospheric turbulence removal methods.
\end{itemize}

\section{Related Works}
 \noindent \textbf{Image Processing methods.} ``Lucky Imaging" techniques \cite{aubailly2009automated,vorontsov2001anisoplanatic} are  the earliest techniques for mitigating atmospheric turbulence. Lucky imaging works by selecting a few good frames from turbulence degraded videos and fusing them to restore the latent sharp image. Anantrasirichai \textit{et al.} \cite{anantrasirichai2013atmospheric} proposed a method to extract regions with fewer degradations from the good frames and fuse them using a dual-tree complex wavelet transform to remove turbulence degradations. Methods based on registration-fusion approaches like \cite{zhu2012removing,hirsch2010efficient,lau2019restoration} were proposed in later years, where a good reference image is initially determined, and the consequent frames are aligned and fused to remove geometric distortions, after which a deconvolution algorithm is applied to remove blur.  Zhu \textit{et al.} \cite{zhu2012removing} use a B-spline registration algorithm and temporal kernel regression to compute a degradation-free image. Lou \textit{et al.}       \cite{lou2013video} proposed a method to reduce deformations between images in a video by applying a temporal smoothing algorithm on video frames sharpened using Sobolev gradient flow. Xie \textit{et al.} \cite{xie2016removing} used a low-rank image as an intermediate reference image and then obtained the sharp image through a variational model.\\
\noindent \textbf{Atmospheric turbulence mitigation using deep networks.}
With the success of CNN-based models in addressing image restoration tasks, a few models have been proposed in the literature to remove atmospheric turbulence. Chak \textit{et al.} \cite{chak2018subsampled} utilize an effective data augmentation algorithm to model turbulence, then filter out the corrupted frames using a sub-sampling algorithm, and finally use a Generative adversarial Network (GAN) to restore the clean image. Recently some methods \cite{yasarla2020learning,lau2021atfacegan,lau2021semi,nair2021confidence} have been proposed for restoring a single face image from a turbulence degraded observations.  Lau \textit{et al.} \cite{lau2019restoration} use the commutative relationship between blur and geometric distortions to generate blur and degradation-free images.  They  further disentangle these images to obtain the clean output. Yasarla \& Patel \cite{yasarla2020learning} propose a  turbulence removal network that estimates blur and degradation priors using Monte Carlo simulations and use them to restore the sharp image. 

\section{Background}
\subsection{Denoising Diffusion Probabilistic models}

Denoising Diffusion Probabilistic Models (DDPMs) \cite{ho2020denoising,sohl2015deep} are a class of generative models that perform image generation through variational inference using a Markovian process with a finite number of timesteps $`T'$. Each step in this Markovian process is a  denoising process. DDPMs consists of two stages -- a forward process and a reverse process. In the forward process, a clean image $y_0$ is sampled, and small Gaussian noises of variance schedules   $\{\beta_1, \dots, \beta_T\}$  are added over $`T'$ timesteps. The overall forward process and each forward step are defined as 
\begin{align}
  q(\by_{0,1....T}) &:=  q(\by_{0})\Pi_{t=1}^{T} q(\by_t | \by_{t-1})\\
  q(\by_t | \by_{t-1}) &= \mathcal{N}\left(\by_t ; \sqrt{1-{\beta}_{t}} \by_{t-1},\sqrt{{\beta}_{t}} \mathbf{I}\right) \\
  &= \sqrt{{\beta}_{t}} \by_{t-1} + \epsilon \sqrt{1-{\beta}_{t}}, \epsilon \sim \mathcal{N}(0, \mathbf{I}),
\end{align}
where $y_t$ and $y_{t-1}$ are noisy samples generated at  timesteps $t$ and $t-1$ and $\{\beta_i\}$ refers to the variance schedules. Since the sum of $t$ zero mean Gaussians with different variance schedules is again a Gaussian, this Markovian process could be effectively represented in terms of the initial data sample $y_0$ and expressed as
\begin{equation}
    \begin{array}{cc}
  q(\by_t | \by_{0}) &:= \mathcal{N}\left(\by_t ; \sqrt{\bar{\alpha}_{t}} \by_0,\left(1-\bar{\alpha}_{t}\right) \mathbf{I}\right) \\
  \vspace{5mm}
  &= \sqrt{\bar{\alpha}_{t}} \by_0 + \epsilon \sqrt{1-\bar{\alpha}_{t}}, \epsilon \sim \mathcal{N}(0, \mathbf{I}),
  \end{array}
  \label{eq:sample_eqn}
\end{equation}
where $\alpha_t =1 - \beta_t$. In the reverse process we define the joint distribution $p_{\theta}(y_{0:T})$ with parameters $\theta$. Similar to the forward process, the reverse process is also a Markovian process which is defined as follows
\begin{align}
    p(y_{T})&:=\mathcal{N}\left(0, \mathbf{I}\right)\\
  p(\by_{t-1} | \by_{t}) &:= \mathcal{N}\left(\by_{t-1} ;  \mu_{\theta}(\by_{t},t),\sqrt{{\beta}_{t}} \mathbf{I}\right).
\end{align}

For optimizing the parameters of the network $\theta$, we minimize the variational lower bound of the the negative log likelihood of the distribution of clean images $y_0$. In this paper, we use the simplified training objective of DDPMs  proposed by Ho \textit{et al.}\cite{ho2020denoising}. For training our model we sample a timestep $t\sim U[1,T]$ and produce the noisy sample corresponding to this timestep using equation(3), defined by 
\begin{equation}
    y_{t}:= \sqrt{\bar{\alpha}_{t}} \by_0 + \epsilon \sqrt{1-\bar{\alpha}_{t}}, \epsilon \sim \mathcal{N}(0, \mathbf{I}).
\end{equation}
The network $p_{\theta}(.)$ predicts the noise $\epsilon$ in this image taking $y_t$ and $t$ as the inputs. The training objective is defined as,
\begin{equation}
 L_{simple} :=E_{t \sim[1, T], \epsilon \sim \mathcal{N}(0, \mathbf{I})}\left[\left\|\epsilon-\epsilon_{\theta}\left(\mathbf{y}_t, t\right)\right\|^{2}\right].\\
\end{equation}



 \begin{figure*}[t!]
	\centering
		\includegraphics[height=8cm,width=.7\textwidth]{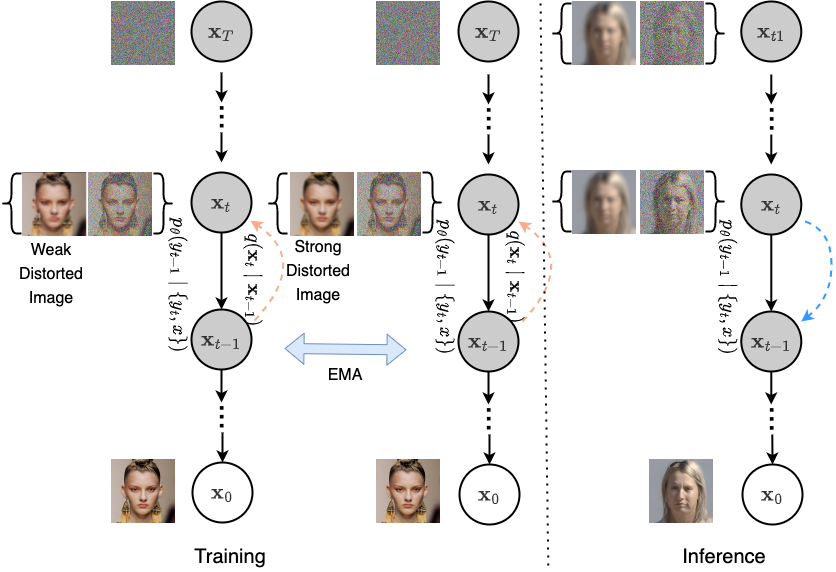}
	\centering
	\caption{An overview of the proposed approach, during the training process, we perform knowledge distillation to transfer class prior information from a network trained for image super-resolution on a large dataset to the network for removing turbulence degradation. During inference, rather than starting from pure Gaussian noise as in normal DDPM, we begin with noised turbulence degraded images for speed-up in inference times. }
	\label{fig:pipeline}
	\vskip -10pt
\end{figure*}
\subsection{Conditional Diffusion Probabilistic Models}

The equations mentioned above are developed for the task of image generation. For utilizing DDPMs for low level vision tasks like image restoration, the conditional distribution of the clean image has to be modelled.
Saharia et al \cite{saharia2021image} have proposed a simple technique for modelling the conditional distribution of clean image  given the corresponding degraded image. In conditional DDPM, the forward process remains the same as that of the unconditional model. Given a clean image  sampled from the dataset, we add  random Gaussian noise based on a randomly sampled timestep $t$. During the reverse process, along with the noisy image and the time $t$, we also pass the degraded image($x$) as input to the neural network. Hence the denoising model is defined by $p_{\theta}(y_t,x,t)$ and the reverse process is defined by
\begin{align}
    p(y_{T})&=\mathcal{N}\left(0, \mathbf{I}\right)\\
  p(\by_{t-1} | \by_{t}, x) &= \mathcal{N}\left(\by_{t-1} ;  \mu_{\theta}(\by_{t},x,t),\sqrt{{\beta}_{t}} \mathbf{I}\right).
  \label{eq:forward_cond}
\end{align}
The mean $\mu_{\theta}(\by_{t},x,t)$ is estimated according to, 
 \begin{align}
     \mu_{\theta}(\by_{t},x,t) = \frac{1}{\sqrt(1-\beta_t)}\left(y_t - \frac{\beta_t}{\sqrt{1-\bar{\alpha}_t}}\epsilon_{\theta}(x,y_t,t)\right).
 \end{align}
 
\subsection{Proposed method}

In this section we detail our proposed method and the training process. For training our model, we perform a multi stage training process as shown in Figure \ref{fig:pipeline}. Given a dataset containing clean facial images, we train a diffusion model $p_{\theta}(.)$ for the task of unconditional generation of facial images. \cite{choi2021ilvr,dhariwal2021diffusion} have already released models trained on large amounts of facial data. Hence we make use of this model as the starting point. Once we have a  model that has learned the distribution of clean facial images, it is easier to adapt those weights to model the conditional distribution  where we aim to restore a slightly degraded image and recover the clean image by conditioning on the distorted image. 

  The same idea holds for the case of two conditional distributions where the diffusion model learns the conditional distribution of clean facial images. For our experiments we choose the conditional distributions as face super-resolution and turbulence since off the shelf super-resolution model trained on a large datasets are already available\cite{dhariwal2021diffusion}. Once we have a model that can perform $(8\times)$ super-resolution(SR), we adapt the model for the stronger degradation of atmospheric turbulence. For making the model robust to the turbulence degradation, we take the model trained for super-resolution and adapt the model to work equally well when the input degraded image is of strong degradation as well as of weak degradation. Let the model trained for the weaker-degradation  be represented by $p_{\phi}(.)$ and the model for restoration of strong turbulence degraded image be denoted by $p_{\delta}(.)$. For optimizing the parameters $\delta$, we sample an $\epsilon \sim \mathcal{N}(0, \mathbf{I})$ and create the noisy sample $y_t$ of the clean image according to Eq.~\ref{eq:sample_eqn}. For each training iteration, the parameters $\delta$ are optimized using the loss function $L_{final}$ defined by 
\begin{equation}
    \begin{array}{cc}
    L_{T}=\left\|\epsilon-\epsilon_{\delta}\left(\mathbf{y}_t,x_{S-turb}, t\right)\right\|^{2}, \epsilon \sim \mathcal{N}(0, \mathbf{I})\\
    L_{S} =  \left\|\epsilon_{\phi}\left(\mathbf{y}_t,x_{W-turb}, t\right)-\epsilon_{\delta}\left(\mathbf{y}_t,x_{S-turb}, t\right)\right\|^{2},\\
 L_{final}= E_{t \sim[1, T]}\left[L_t + \gamma L_S\right].
 \end{array}
\end{equation}
Here the term $L_S$ ensures that the model focuses on reconstructing the same face regardless of the distortion nature. Once the parameters $\delta$  are updated by optimizing the loss function $L_{final}$, we perform an exponential moving average (EMA) based updating of the weights of the SR model. Please not that the weights $\phi$ are not updated by optimizing the loss function. The EMA-based weight update for the weights $\phi$ using the estimated weights $\delta$ is according to 
\begin{equation}
    \phi = \gamma_1 \phi + (1-\gamma_1)\delta. 
\end{equation}
Hence adapting the model $p_{\phi}(.)$ for weaker degradation. 

\noindent\textbf{Efficient inference for turbulence removal:} The usual inference process for diffusion models is very time consuming. However we observed that for the task of conditional image generation, this is not the case and the model creates good results even for very low number of timesteps ranging $T=40-50$. Furthermore, we noted that the initial steps in DDPMs learn the coarse features. But these features are already present in the turbulence distorted image. Hence rather than starting from pure Gaussian noise, the inference process could be started from the noised turbulence distorted image. Let $x$ denote the turbulence distorted input image, and $y_t$ the sampled inference image after $t$ forward steps in diffusion. As mentioned in Eq.~ \ref{eq:forward_cond}, the reverse step starts with Gaussian noise at $t=T$. We start the diffusion process from a time $t=t_1$ than $T$. We explicitly assign $y_t1$ by
\begin{equation}
    y_{t1} =q(y_t|y_0), t=t_1.
\end{equation}
There also exist a hidden advantage of this efficient sampling strategy. Atmospheric turbulence mitigation by itself is a highly ill-posed problem. Since the diffusion process is also stochastic, it tends to increase the ill-posedness by sampling a face that is close to the distorted face during inference. Starting from the turbulence distorted images fixes the coarse features in the image to be sampled hence reducing the ill-posedness.


\begin{figure*}[htbp]
    \centering
    \begin{subfigure}[t]{0.120\linewidth}
      \captionsetup{justification=centering, labelformat=empty, font=scriptsize}
      \includegraphics[width=1\linewidth]{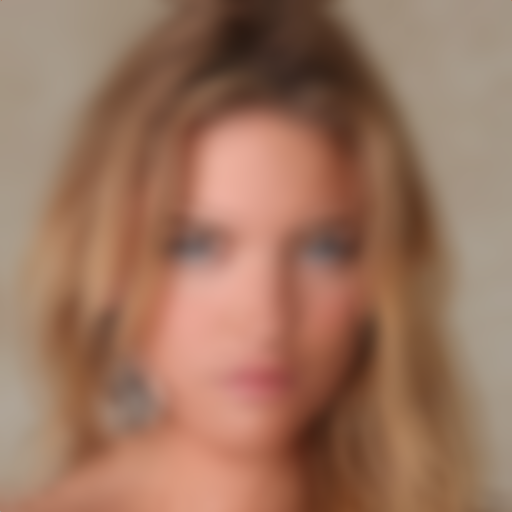}
      \includegraphics[width=1\linewidth]{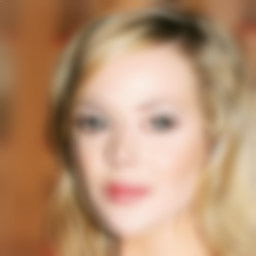}
      \includegraphics[width=1\linewidth]{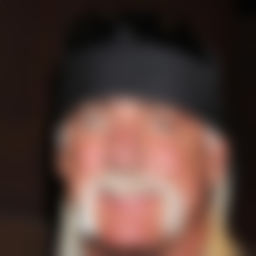}
      \caption{Distorted}
    \end{subfigure}
    \begin{subfigure}[t]{0.120\linewidth}
      \captionsetup{justification=centering, labelformat=empty, font=scriptsize}
      \includegraphics[width=1\linewidth]{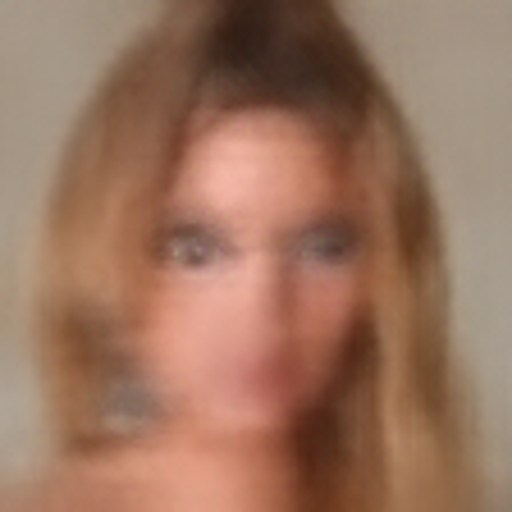}
      \includegraphics[width=1\linewidth]{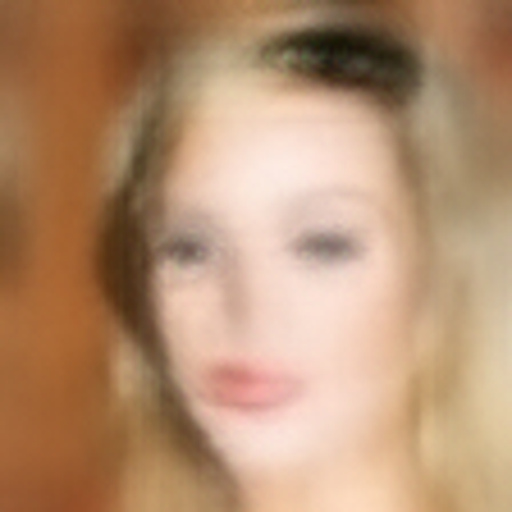}
      \includegraphics[width=1\linewidth]{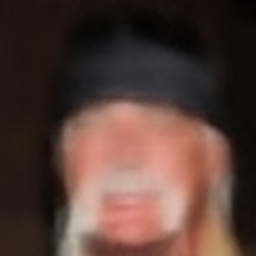}
      \caption{ATFaceGAN\cite{lau2020atfacegan}}
    \end{subfigure}
    \begin{subfigure}[t]{0.120\linewidth}
      \captionsetup{justification=centering, labelformat=empty, font=scriptsize}
      \includegraphics[width=1\linewidth]{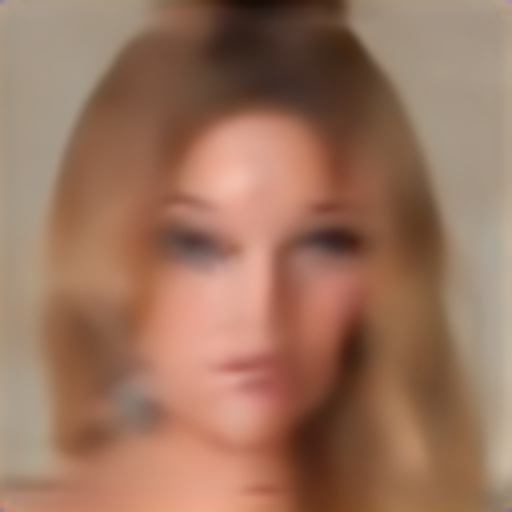}
      \includegraphics[width=1\linewidth]{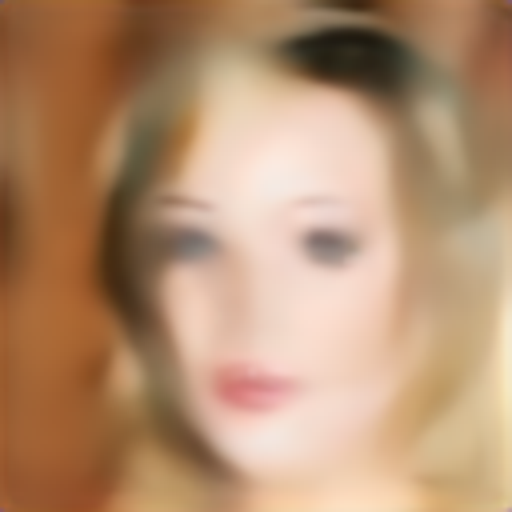}
      \includegraphics[width=1\linewidth]{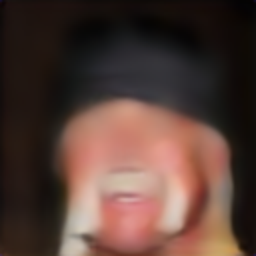}
      \caption{ATNet\cite{yasarla2020learning}}
    \end{subfigure}
    \begin{subfigure}[t]{0.120\linewidth}
      \captionsetup{justification=centering, labelformat=empty, font=scriptsize}
      \includegraphics[width=1\linewidth]{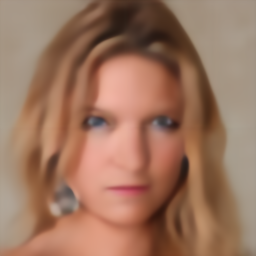}
      \includegraphics[width=1\linewidth]{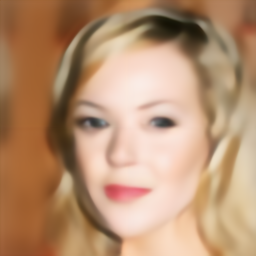}
      \includegraphics[width=1\linewidth]{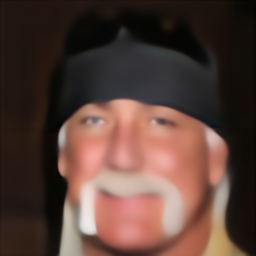}
      \caption{MPRNet\cite{zamir2021multi}}
    \end{subfigure}
    \begin{subfigure}[t]{0.120\linewidth}
      \captionsetup{justification=centering, labelformat=empty, font=scriptsize}
      \includegraphics[width=1\linewidth]{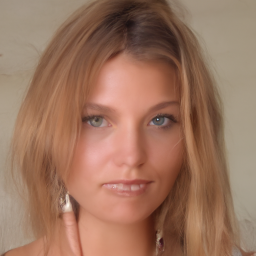}
      \includegraphics[width=1\linewidth]{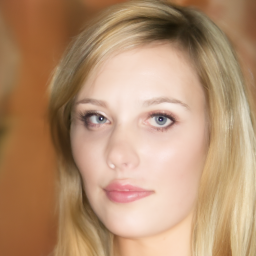}
      \includegraphics[width=1\linewidth]{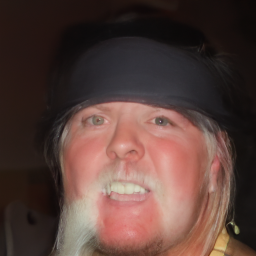}
      \caption{ILVR\cite{choi2021ilvr}}
    \end{subfigure}
    \begin{subfigure}[t]{0.120\linewidth}
      \captionsetup{justification=centering, labelformat=empty, font=scriptsize}
      \includegraphics[width=1\linewidth]{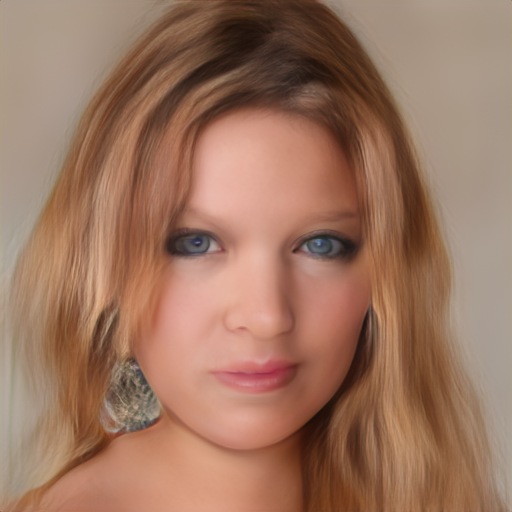}
      \includegraphics[width=1\linewidth]{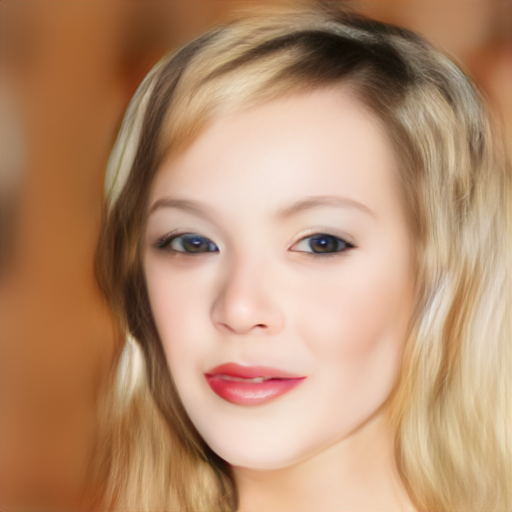}
      \includegraphics[width=1\linewidth]{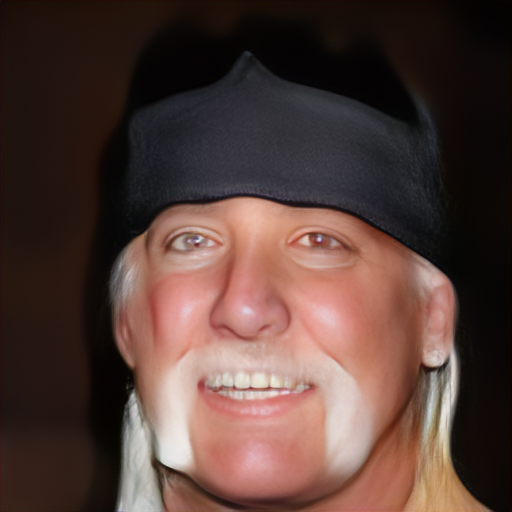}
      \caption{LTTGAN\cite{mei2021ltt}}
    \end{subfigure}
    \begin{subfigure}[t]{0.120\linewidth}
      \captionsetup{justification=centering, labelformat=empty, font=scriptsize}
      \includegraphics[width=1\linewidth]{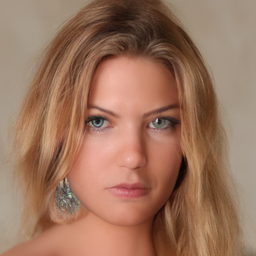}
      \includegraphics[width=1\linewidth]{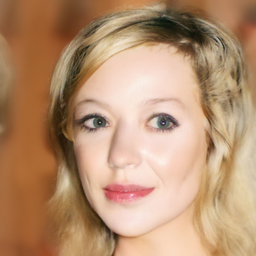}
      \includegraphics[width=1\linewidth]{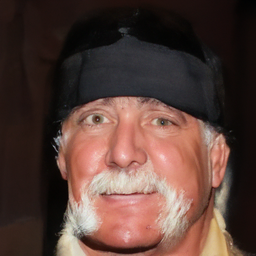}
      \caption{OURS}
    \end{subfigure}
    \begin{subfigure}[t]{0.120\linewidth}
      \captionsetup{justification=centering, labelformat=empty, font=scriptsize}
      \includegraphics[width=1\linewidth]{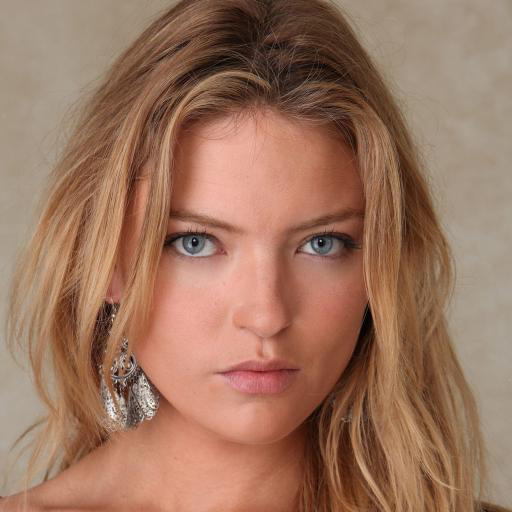}
      \includegraphics[width=1\linewidth]{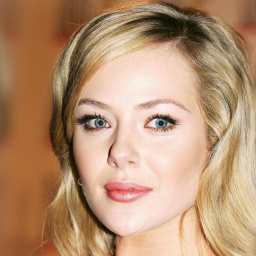}
      \includegraphics[width=1\linewidth]{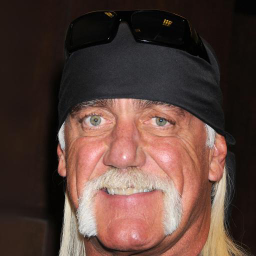}
      \caption{GT}
    \end{subfigure}
    \vspace{-3mm}    \caption{Qualitative comparisons with single image atmospheric turbulence mitigation methods (ATFaceGAN, TDRN) as well as other generative models (ILVR,GFPGAN) on CelebA dataset\cite{liu2015faceattributes}. }
    \label{fig:facesr}
    \vspace{-5mm}
  \end{figure*}

\section{Experiments}
For quantitative evaluations, we utilize two different type of metrics, namely image quality-based metrics and facial recognition-based metrics. For evaluating the reconstructed image based on the clean target, we use Peak Signal to Noise Ratio (PSNR) and Structural Similarity Index (SSIM). PSNR might not always capture the quality of reconstructed faces since blurry images could have higher PSNR than clean images \cite{zhang2018perceptual}. Hence, for evaluating the the quality of the outputs generated  by our model, we also utilize other perceptual and no-reference quality metrics. Specifically, we use the Naturalness Image Quality Evaluator (NIQE score) \cite{mittal2012making} and the LPIPS distance \cite{zhang2018perceptual} for evaluating the naturalness of the generated image and measuring the perceptual similarity in the  feature space. To assess how close the images are to the distribution of clean facial images, we utilize Fréchet Inception Distance (FID score) \cite{heusel2017gans}. As for the facial recognition-based metrics, we use face recognition score calculated using ArcFace \cite{deng2019arcface}.  We use three different recognition scores for comparing the performance -- Top-1, Top-3 and Top-5. The Top-K score refers to the actual identity being present in the K-nearest faces  when the distance between the features of the measured face and the faces in the gallery set is computed.
\subsection{Training details}
We set the number of training steps as $T=1000$ and the number of inference steps as  $T=60$. The inference time is reduced by applying the time rescaling strategy mentioned in \cite{nichol2021improved} and $t_1=30$ for all our experiments. $\gamma=0.01$ and $\gamma_1=0.9909$ for all our experiments.

\subsection{Training dataset}
\label{sec:dataset}
Various methods have been proposed in the literature for simulating atmospheric turbulence degradation\cite{nair2022comparison, schwartzman2017turbulence,lau2020atfacegan,mei2021ltt,mao2021accelerating}. All of these models aim to generate a spatially varying Point Spread Functions (PSFs) and distort each pixel in a clean image. In our work, we follow the model used in  \textit{Mei and Patel} \cite{mei2021ltt} which is based on  the observation model in Eq.~\ref{eq: turb}. To create a spatially varying PSF and a geometric distortion at the same time, we use an elastic transformation along with a  blur augmentation.  Finally, we add an additive white Gaussian noise with a standard deviation $1e^{-4}$. For training we use 35,000 randomly sampled images from the FFHQ dataset \cite{karras2019style}.


\subsection{Testing Datasets} 
 We evaluate our method by performing experiments on one synthetic face dataset, and two real-world datasets (LRFID dataset and BRIAR dataset) consisting of facial images degraded by atmospheric turbulence. We use the CelebA dataset \cite{liu2015faceattributes} and apply elastic blur \cite{mei2021ltt} to create a synthetic test set. In total, we select 100 identities from the CelebA dataset. For each image, we resize the face to size $256 \times 256$ and introduce the degradation.  For comparing the performance in real-world scenarios, we use the LRFID dataset \cite{miller2019data} and the BRIAR dataset.  The LRFID dataset consists of 89 videos in different atmospheric conditions corresponding to 89 different individuals. We extract one image frame randomly from each video and crop it in order to generate the real-world test set. The LRFID dataset also has clean images of the same individuals captured in close range conditions. Since we do not have the clean target pairs of these videos, we utilize clear images of these individuals taken in outdoor conditions as reference and perform facial recognition using ArcFace \cite{deng2019arcface} as well as other perceptual quality metrics to compare the performance of different networks. The BRIAR dataset is another real-world dataset consisting of high-resolution video data of 343 individuals. The videos are shot at different ranges from the individuals varying from 100m to 500m. For our experiments, we utilize a subset of the BRIAR dataset corresponding to ranges 200m and 400m. In total there are 130 identities with data in these ranges. We extract a frame from one video per identity. Since the resolution of the captured frames are quite high, we utilize retinaface face detection algorithm \cite{deng2019retinaface} to extract faces from the center portion of each image. The testset that we use consists of a total of 50 identities.

\subsection{Results}
\begin{figure*}[tbp]
    \centering
    \begin{subfigure}[t]{0.120\linewidth}
      \captionsetup{justification=centering, labelformat=empty, font=scriptsize}
      \includegraphics[width=1\linewidth]{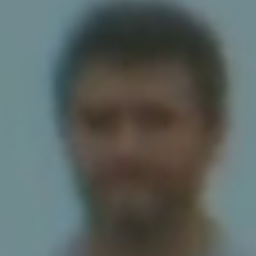}
      \includegraphics[width=1\linewidth]{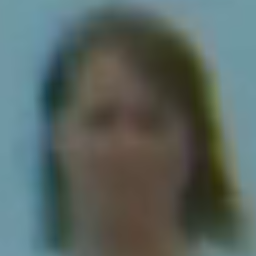}
      \includegraphics[width=1\linewidth]{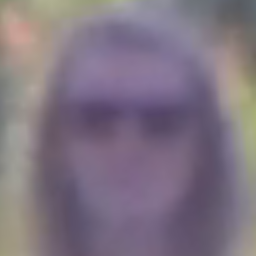}
      \caption{Distorted}
    \end{subfigure}
    \begin{subfigure}[t]{0.120\linewidth}
      \captionsetup{justification=centering, labelformat=empty, font=scriptsize}
       \includegraphics[width=1\linewidth]{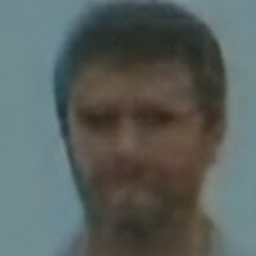}
      \includegraphics[width=1\linewidth]{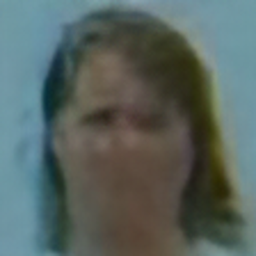}
      \includegraphics[width=1\linewidth]{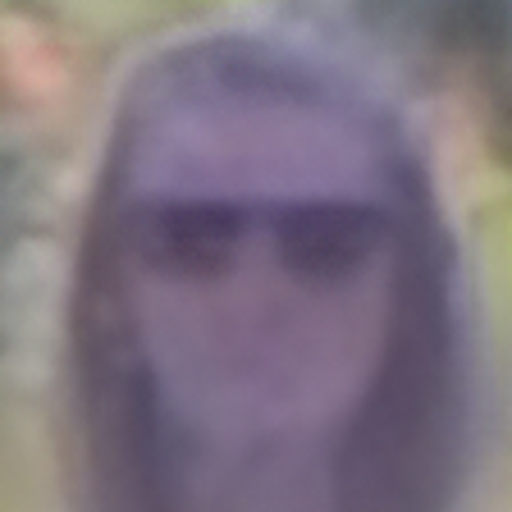}
      \caption{ATFaceGAN\cite{lau2020atfacegan}}
    \end{subfigure}
    \begin{subfigure}[t]{0.120\linewidth}
      \captionsetup{justification=centering, labelformat=empty, font=scriptsize}
      \includegraphics[width=1\linewidth]{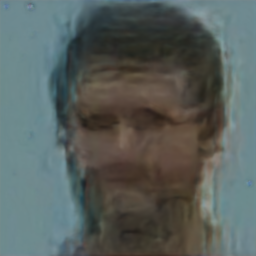}
      \includegraphics[width=1\linewidth]{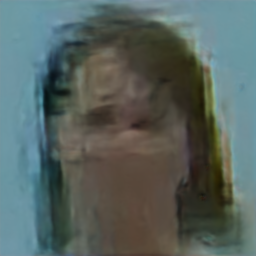}
      \includegraphics[width=1\linewidth]{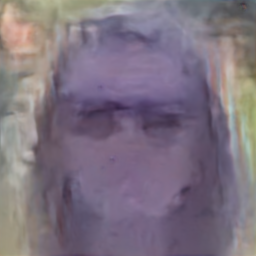}
      \caption{ATNet\cite{yasarla2021learning}}
    \end{subfigure}
    \begin{subfigure}[t]{0.120\linewidth}
      \captionsetup{justification=centering, labelformat=empty, font=scriptsize}
      \includegraphics[width=1\linewidth]{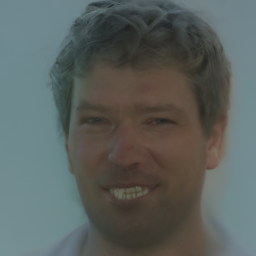}
      \includegraphics[width=1\linewidth]{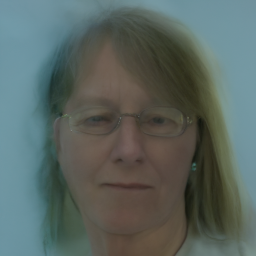}
      \includegraphics[width=1\linewidth]{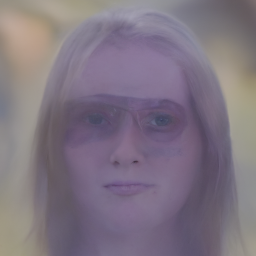}
      \caption{ILVR\cite{choi2021ilvr}}
    \end{subfigure}
    \begin{subfigure}[t]{0.120\linewidth}
      \captionsetup{justification=centering, labelformat=empty, font=scriptsize}
      \includegraphics[width=1\linewidth]{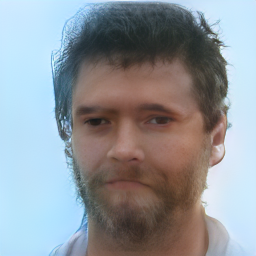}
      \includegraphics[width=1\linewidth]{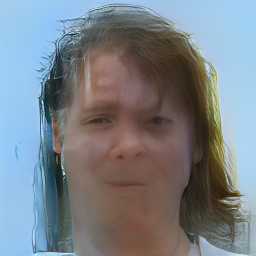}
      \includegraphics[width=1\linewidth]{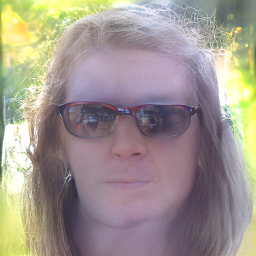}
      \caption{GFPGAN\cite{wang2021towards}}
    \end{subfigure}
    \begin{subfigure}[t]{0.120\linewidth}
      \captionsetup{justification=centering, labelformat=empty, font=scriptsize}
      \includegraphics[width=1\linewidth]{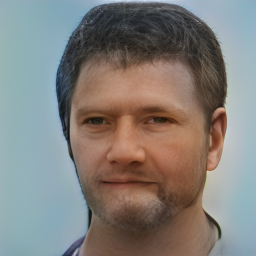}
      \includegraphics[width=1\linewidth]{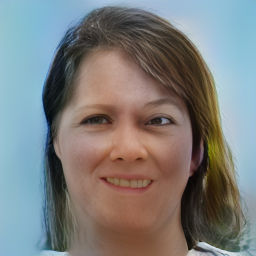}
      \includegraphics[width=1\linewidth]{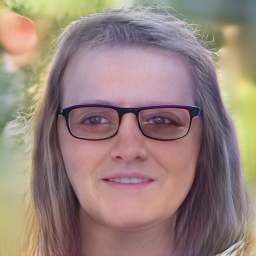}
      \caption{LTTGAN\cite{mei2021ltt}}
    \end{subfigure}
    \begin{subfigure}[t]{0.120\linewidth}
      \captionsetup{justification=centering, labelformat=empty, font=scriptsize}
      \includegraphics[width=1\linewidth]{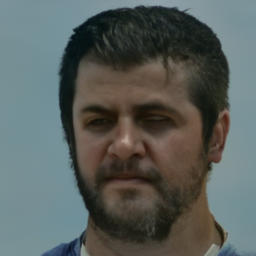}
      \includegraphics[width=1\linewidth]{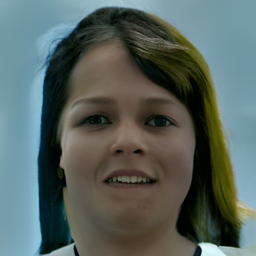}
      \includegraphics[width=1\linewidth]{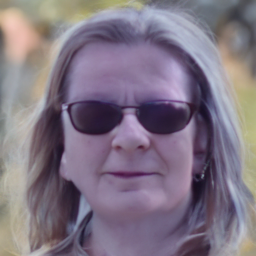}
      \caption{OURS}
    \end{subfigure}
    \begin{subfigure}[t]{0.120\linewidth}
      \captionsetup{justification=centering, labelformat=empty, font=scriptsize}
      \includegraphics[width=1\linewidth]{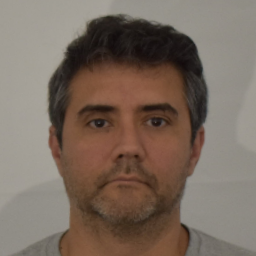}
      \includegraphics[width=1\linewidth]{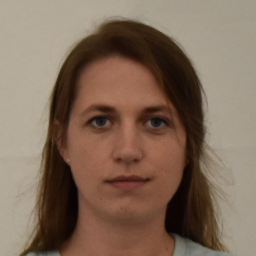}
      \includegraphics[width=1\linewidth]{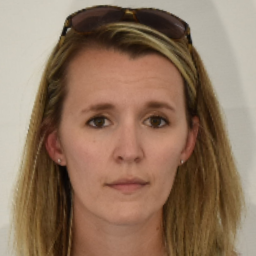}
      \caption{Gallery}
    \end{subfigure}
    \vspace{-3mm}    \caption{Qualitative comparisons with single image atmospheric turbulence mitigation methods on the BRIAR dataset. }
    \label{fig:facebriar}
    \vspace{-0.5cm}
  \end{figure*}

We qualitatively and quantitatively compare our method with various deep networks for atmospheric turbulence removal and face restoration. We utilize CNN-based models, GAN-based models and diffusion models. For CNN-based atmospheric turbulence mitigation methods, we use MPRNet \cite{zamir2021multi} and ATNet \cite{yasarla2021learning}. We also compare our method with face enhancement networks built upon generative models. Here, we choose two different varieties of generative models, specifically GAN-based method for atmospheric turbulence removal ATFaceGAN \cite{lau2020atfacegan} as well as two GAN inversion-based models \cite{mei2021ltt, wang2021towards} trained for removing atmospheric turbulence. Finally, we compare our method with one diffusion model for image restoration \cite{choi2021ilvr}. All the above mentioned methods are retrained with our dataset for fair comparisons. The quantitative analysis can be found in Table~\ref{table:quant} and the qualitative results on the CelebA dataset \cite{liu2015faceattributes}, the LRFID dataset \cite{miller2019data} and the BRIAR dataset can be found in Figure \ref{fig:facesr}, \ref{fig:facelrfid} and \ref{fig:facebriar}, respectively.



\subsection{Qualitative Analysis}

\noindent\textbf{Synthetic dataset evaluation:} The quantitative results for celebA dataset \cite{liu2015faceattributes} are presented in Table \ref{table:quant}. The values marked in dark green colour for each metric denotes the best value among all comparison methods and the ones in light green colour denote the second best value. As can be seen from Table \ref{table:quant}, the highest PSNR  and SSIM are achieved by \cite{zamir2021multi}. This is due to the fact that CNNs are generally trained to optimize the Mean square error(MSE) loss between the network prediction and the clean target. Reducing the MSE loss is proportional to increasing the PSNR. Hence an image with a higher PSNR compared to the clean target could be in fact more distorted than an image with lower PSNR. This could be clearly seen from Figure \ref{fig:facesr}.  Hence we make use of the no reference metric NIQE that estimates the naturalness of the predicted output. Lower the NIQE score, more natural is the image. We can see that GFPGAN \cite{wang2021towards} generates the outputs with the lowest NIQE score. Our method generates outputs with the next lowest NIQE score values. The next metric we utilize is the LPIPS \cite{zhang2018perceptual} score which compares the perceptual similarity of the predictions from the networks and the set of clean images is the LPIPS score. As we can see, the GAN-inversion based method LTTGAN \cite{mei2021ltt} has the best LPIPS distance and our method produces outputs with the next best LPIPS distance. 
\begin{table}[t!]
	\begin{center}
		\resizebox{0.47\textwidth}{!}{
			\begin{tabular}{|c | c c c c c | }
			 \hline
			  \multirow{1}{*}{ Dataset }&   \multicolumn{5}{|c|}{CelebA dataset\cite{liu2015faceattributes}} \\
				\hline
				Metric&PSNR$(\uparrow)$&SSIM$(\uparrow)$&NIQE$(\downarrow)$&LPIPS$(\downarrow)$&FID$(\downarrow)$\\		\hline
				GT&inf&1.000&5.750&0&0 \\
				degraded&22.76&0.6517&21.35&0.4642&146.72\\
				\hline
				    &\multicolumn{5}{|c|}{CNN based models}\\
				MPRNET\cite{yasarla2021learning}&\cellcolor{green!50}24.28&\cellcolor{green!50}0.6951&10.29&0.3885&121.37\\	
				ATNet\cite{yasarla2021learning}&21.88&0.6271&13.20&0.5065&169.52\\
								\hline
				       &\multicolumn{5}{|c|}{GAN based models}\\
		ATFaceGAN\cite{lau2020atfacegan}&22.11&0.5821&11.49&0.4887&182.18\\
				GFPGAN\cite{wang2021towards}&21.74&0.5925&\cellcolor{green!50}6.083&0.3341&\cellcolor{green!25}75.338\\
				LTTGAN\cite{mei2021ltt}&\cellcolor{green!25}22.77& \cellcolor{green!25}0.6458&7.884&\cellcolor{green!50}0.3192&82.182\\
				\hline
				      & \multicolumn{5}{|c|}{Diffusion models}\\	
				ILVR\cite{choi2021ilvr}&21.40&0.5916&7.762&0.3689&76.702\\
				OURS&22.41&0.6262&\cellcolor{green!25}7.186&\cellcolor{green!25}0.3223&\cellcolor{green!50}73.127\\
				\hline
			\end{tabular}
		}
					\caption{Quantitative results on synthetic turbulence degraded images prepared using CelebA dataset\cite{liu2015faceattributes} 
			\label{table:quant}}
	\end{center}
	\vspace{-0.8cm}
\end{table}

To estimate how the predicted distribution of the networks vary from the actual distribution of clean facial images, we compare the predictions from the networks in terms of the FID score. The generative models create outputs with much better FID score when compared to non-generative models as can be seen from Table~\ref{table:quant}. Our method generates outputs with the best FID score and GFPGAN generates outputs with the second best FID score. The qualitative results on the CelebA dataset can be seen in Figure \ref{fig:facesr}. As we can see the CNN-based methods ATNet \cite{yasarla2021learning} and MPRNet \cite{zamir2021multi} and GAN restoration method ATFaceGAN \cite{lau2020atfacegan} remove the turbulence distortion  to some extent but fail to reconstruct relevant facial features like eyes, nose and hair and overly smoothen the facial features. Compared to these techniques, the generative modelling based facial restoration techniques LTTGAN \cite{mei2021ltt} and ILVR \cite{choi2021ilvr} reconstruct images with much better facial features. ILVR \cite{choi2021ilvr} reconstructs faces with good features but  creates faces that are much different from the original faces.  In case of LTTGAN \cite{mei2021ltt}, although much better faces are reconstructed, the reconstructed image differs in multiple factors like eye color, shape of face which can be clearly seen from the degraded images. In contrast our method is able to reconstruct more accurate faces.

\begin{table*}[t!]
	\begin{center}
		\resizebox{\textwidth}{!}{
			\begin{tabular}{|c | c c c c c c| c c c c c c|}
			 \hline
			  \multirow{1}{*}{ Dataset }&   \multicolumn{6}{|c|}{BRIAR dataset} & \multicolumn{6}{|c|}{LRFID dataset\cite{miller2019data}}\\
				\hline
				Metric&NIQE$(\downarrow)$&LPIPS$(\downarrow)$&FID$(\downarrow)$&Top-1$(\uparrow)$&Top-3$(\uparrow)$&Top-5$(\uparrow)$&NIQE$(\downarrow)$&LPIPS$(\downarrow)$&FID$(\downarrow)$&Top-1$(\uparrow)$&Top-3$(\uparrow)$&Top-5$(\uparrow)$\\		\hline
				GT&5.867&0&0&100.0&100.00&100.0 & 5.729&0&0&100.0&100.0&100.0\\
				degraded&16.68&0.7754&317.82&26.0&38.0&46.0&12.90&0.6293&195.71&35.3&62.2&71.2\\
				\hline
				    &\multicolumn{12}{|c|}{CNN based models}\\
				MPRNET\cite{yasarla2021learning}&11.76&0.7051&220.73&24.0&46.0&64.0& 11.13&0.5755&176.41&34.1&64.6&74.4\\	
				ATNet\cite{yasarla2021learning}&8.506&0.7265&314.82&14.0&28.0&38.0& 12.24&0.6128&202.45&36.5&64.6&74.4\\
								\hline
				       &\multicolumn{12}{|c|}{GAN based models}\\
		ATFaceGAN\cite{lau2020atfacegan}&10.66&0.7459&260.25&22.0&38.0&50.0& 9.434&0.6300&169.60&47.5&65.8&82.3\\
				GFPGAN\cite{wang2021towards}&6.854&0.6414&204.76&\cellcolor{green!25}26.0&\cellcolor{green!50}58.0&60.0&\cellcolor{green!25}7.918&0.5587&124.55&57.3&79.2&85.3\\
				LTTGAN\cite{mei2021ltt}&\cellcolor{green!50}5.500&\cellcolor{green!50}0.5969& \cellcolor{green!50}150.24&20.0&54.0&\cellcolor{green!25}62.0&7.970&\cellcolor{green!50}0.4803&\cellcolor{green!25}119.23&\cellcolor{green!25}58.5&\cellcolor{green!25}81.7&\cellcolor{green!25}85.3\\
				\hline
				      & \multicolumn{12}{|c|}{Diffusion models}\\	
				ILVR\cite{choi2021ilvr}&8.145&0.6523&167.78&22.0&44.0&56.0&12.068&0.5661&161.38&31.7&59.7&67.0\\
				OURS&\cellcolor{green!25}6.283&\cellcolor{green!25}0.6368&\cellcolor{green!25}152.57&\cellcolor{green!50}32.0&\cellcolor{green!25}56.0&\cellcolor{green!50}66.0&\cellcolor{green!50}7.576&\cellcolor{green!25}0.5234&\cellcolor{green!50}112.64&\cellcolor{green!50}62.2&\cellcolor{green!50}81.7&\cellcolor{green!50}87.8\\
				\hline
			\end{tabular}
		}
					\caption{Quantitative results on on real world turbulence degraded datasets: BRIAR dataset and LRFID dataset \cite{miller2019data}
			\label{table:quant}}
	\end{center}
	\vspace{-0.8cm}
\end{table*}

\noindent\textbf{Results on the LRFID dataset \cite{miller2019data}:} Since the ground truth targets for the LRFID dataset \cite{miller2019data} are not available, we compare the performance of the networks qualitatively using no reference metrics, perceptual quality metrics as well as facial recognition accuracy. By comparing the naturalness of the generated outputs using the NIQE score, we can see that the generative methods work much better than the CNN-based models. Our method creates the most natural images and we are able to obtain a NIQE score of $1.094$ above the previous best method. Based on the LPIPS metric, the GAN inversion method LTTGAN \cite{mei2021ltt} works the best and our model performs the second best. To evaluate  how the generated images are close to real-world facial images, we use the FID score.  Regarding the FID score, it can be seen from Table \ref{table:quant} that our model works much better than the other models with an FID score $13$ below the next best method. Also we can see that the generative models generate images with much better facial features. To further analyze the quality of the restored images, we compare the facial recognition accuracy of the restored images using Arcface facial recognition  framework \cite{deng2019arcface} with resnet34 backbone.  Based on the facial recognition scores, we can see that the Top-1 facial recognition accuracy is obtained from the images restored by LTTGAN\cite{mei2021ltt}.  We obtain the second best facial recognition accuracy, which is just $1.2$ below LTTGAN. Our method is able to obtain much better facial recognition accuracies compared to the other methods. The results on the real world LRFID dataset \cite{miller2019data} can be seen in Figure \ref{fig:facelrfid}.  As we can see from this figure,  ATNet \cite{yasarla2021learning} is able to remove small amount of distortions, but is not able to reconstruct facial features well. Compared to this ATNet works a bit better and is able to reconstruct faces with much better features.  The diffusion-based model \cite{choi2021ilvr} is primarily defined for super-resolution and fails on real-world turbulence distorted images. As we can see from Figure \ref{fig:facelrfid}, GFPGAN \cite{wang2021towards} and LTTGAN \cite{mei2021ltt} slightly change the color of the reconstructed images. LTTGAN \cite{mei2021ltt} also creates vertical line artifacts on top of the image. In contrast, our model doesn't cause a colour shift on the reconstructed image and is able to remove turbulence degradation.
\begin{figure*}[!htb]
 \centering
    \begin{subfigure}[t]{0.16\linewidth}
      \captionsetup{justification=centering, labelformat=empty, font=scriptsize}
    \includegraphics[width=1\linewidth]{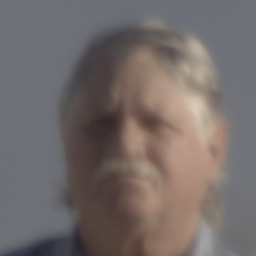}
      \caption{(a)}
    \end{subfigure}
    \begin{subfigure}[t]{0.16\linewidth}
      \captionsetup{justification=centering, labelformat=empty, font=scriptsize}
    \includegraphics[width=1\linewidth]{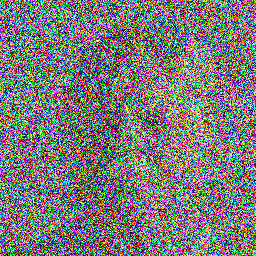}
    
      \caption{(b)}
    \end{subfigure}
    \begin{subfigure}[t]{0.16\linewidth}
      \captionsetup{justification=centering, labelformat=empty, font=scriptsize}
    \includegraphics[width=1\linewidth]{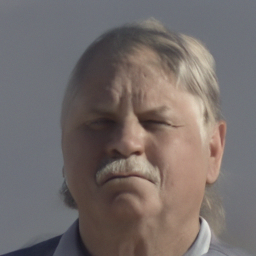}
      \caption{(c)}
    \end{subfigure}
  \begin{subfigure}[t]{0.16\linewidth}
      \captionsetup{justification=centering, labelformat=empty, font=scriptsize}
      \includegraphics[width=1\linewidth]{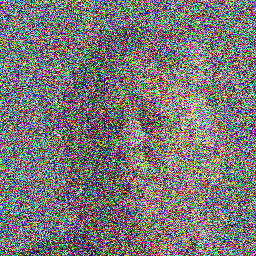}
      \caption{(d)}
    \end{subfigure}
    \begin{subfigure}[t]{0.16\linewidth}
      \captionsetup{justification=centering, labelformat=empty, font=scriptsize}
      \includegraphics[width=1\linewidth]{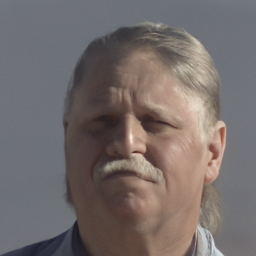}

      \caption{(e)}
    \end{subfigure}
    \begin{subfigure}[t]{0.16\linewidth}
      \captionsetup{justification=centering, labelformat=empty, font=scriptsize}
    \includegraphics[width=1\linewidth]{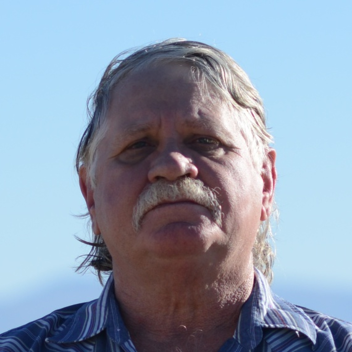}
      \caption{(f)}
    \end{subfigure}
    \vskip-10pt\caption{(a) Input turbulence-degraded image. (b) Generated image corresponding to $T=20$ starting from Gaussian noise. (c) Generated image corresponding to $T=0$ starting from Gaussian noise. (d) Turbulence image added with Gaussian noise corresponding to $T=20$. (e) Final reconstructed Image starting from turbulence degraded image. (f) Clean Image.}
        \label{fig:ablT}
    \vspace{-4mm}
  \end{figure*}




\section{Results on the BRIAR dataset}

Figure \ref{fig:facebriar} shows the qualitative results on the BRIAR dataset. As we can see the images in the BRIAR dataset are very much distorted hence making it very difficult to restore these images. Because of the high level of distortion, most methods fail on the BRIAR dataset. The existing generative modelling-based methods fail to reconstruct the relevant facial features in the image. Our method although cannot reconstruct the exact identity properly, is able to reconstruct a structured face that could be retrieved from the given input image.

\section{Ablation Studies}

\subsection{Effect of Progressive Training:-}
\begin{figure}[tp]
    \centering
    \begin{subfigure}[t]{0.240\linewidth}
      \captionsetup{justification=centering, labelformat=empty, font=scriptsize}
      \includegraphics[width=1\linewidth]{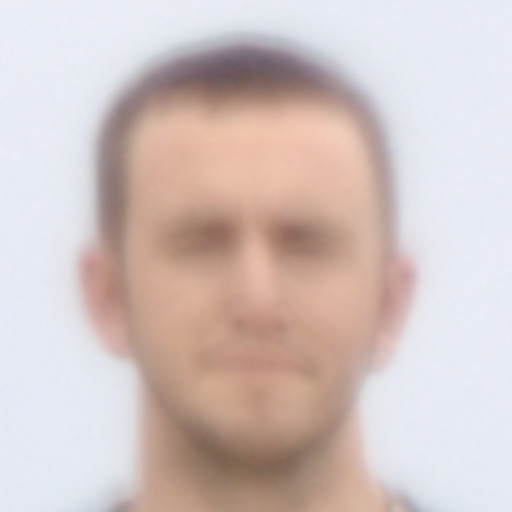}
      \includegraphics[width=1\linewidth]{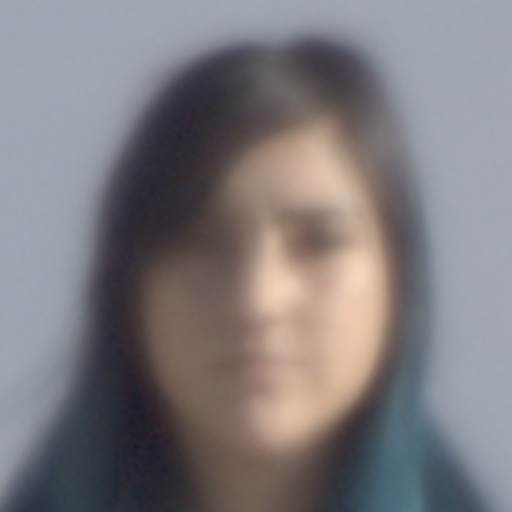}
      \caption{Distorted}
    \end{subfigure}
    \begin{subfigure}[t]{0.240\linewidth}
      \captionsetup{justification=centering, labelformat=empty, font=scriptsize}
\includegraphics[width=1\linewidth]{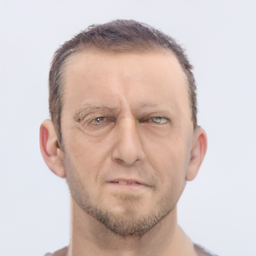}
      \includegraphics[width=1\linewidth]{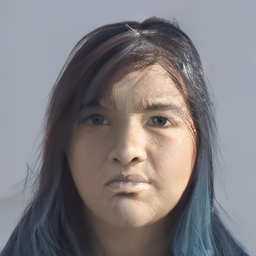}
      \caption{Without PT}
    \end{subfigure}
    \begin{subfigure}[t]{0.240\linewidth}
      \captionsetup{justification=centering, labelformat=empty, font=scriptsize}
\includegraphics[width=1\linewidth]{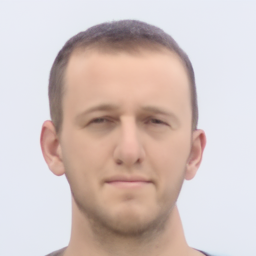}
      \includegraphics[width=1\linewidth]{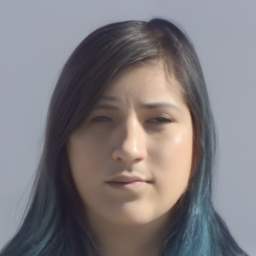}

      \caption{OURS}
    \end{subfigure}
    \begin{subfigure}[t]{0.240\linewidth}
      \captionsetup{justification=centering, labelformat=empty, font=scriptsize}
\includegraphics[width=1\linewidth]{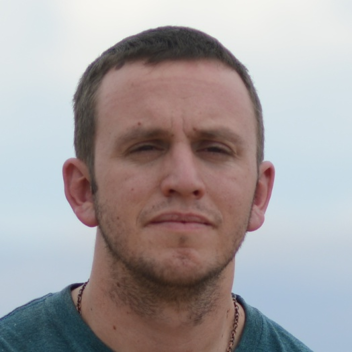}
      \includegraphics[width=1\linewidth]{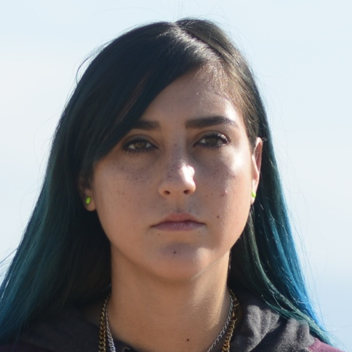}
      \caption{GT}
    \end{subfigure}
   
    \vspace{-3mm}    \caption{Illustration of improvement obtained by the proposed progressive training method.}
    \label{fig:faceabl}
  \end{figure}

To compare the performance improvements obtained by the newly proposed way of training, we retrain the network in a simple manner by directly conditioning the diffusion model with the turbulence distorted image. We train the model for twice the number of iterations required for training our base network. We then test on the real-world LRFID dataset\cite{miller2019data} since it contains images much different than those used for training. A visualization of improvement brought by the progressive training procedure is shown in Fig.~\ref{fig:faceabl}. The qualitative comparisons for both the models can be seen from Table~\ref{table:abla}. We can see that there is a significant boost in performance with our method of training.
\subsection{Effect of efficient sampling} 
\begin{figure}[t!]
	\centering
		\includegraphics[width=0.45\textwidth]{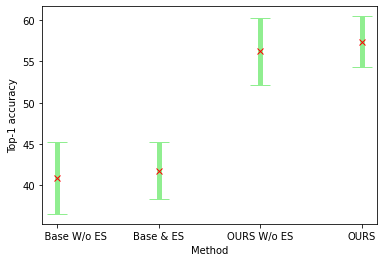}
	\centering
\vskip -10pt	\caption{Error bar plot illustrating the effect of efficient sampling.}
	\label{fig:error}
\vspace{-4mm}
\end{figure}

Figure \ref{fig:ablT} shows an illustration of our efficient sampling method. As can be seen from Fig \ref{fig:ablT}, at $T=20$ the generated image looks like a course face with noise added to it. As we can see from Fig \ref{fig:ablT} (d) the noised turbulence degraded image looks very similar to this. Figures \ref{fig:ablT} (c) and \ref{fig:ablT} (e)  show a visualization of the final reconstructed outputs using these starting initializations. As we can see, qualitatively our method is able to preserve the identity pretty well. Moreover, we create 10 samples per inference image and plot the error bar for Top-1 facial verification accuracies in Fig.\ref{fig:error}. The facial verification accuracies are lower when the efficient sampling technique is used. Hence validating our claims.
\begin{table}[t!]
	\begin{center}
		\resizebox{0.47\textwidth}{!}{
			\begin{tabular}{|c | c c c c c  | }
				\hline
				Metric&Top-1$(\uparrow)$&Top-3$(\uparrow)$&Top-5$(\uparrow)$&NIQE$(\downarrow)$&LPIPS$(\downarrow)$\\	
                \hline
				Without PT&48.7&74.2&79.3&6.985&0.5358\\
				OURS&54.8&84.1&86.5&6.824&0.5255\\
				\hline
			\end{tabular}
		}
					\caption{An ablation study with and without the proposed progressive training on the LRFID dataset \cite{miller2019data}.
			\label{table:abla}}
	\end{center}
	\vspace{-0.8cm}
\end{table}

\section{Conclusion}
In this paper, we proposed an effective generative modelling based solution for reconstructing facial images degraded by atmospheric turbulence. A new strategy is proposed for training diffusion models for inverse problems. We make use of a model trained for super resolution on a large facial dataset and adapt it to learn the conditional distribution of clean images given turbulence degraded images. We also introduce a new efficient sampling strategy to reduce the inference time and stabilize the DDPM outputs  during inference. We achieve the state-of-the-art results on one synthetic and two real-world datasets. Furthermore, we perform extensive ablation analysis to show the improvement obtained by different parts of our approach. 
 
\section{Acknowledgement}

This research is based upon work supported in part by the Ofﬁce of the Director of National Intelligence (ODNI), Intelligence Advanced Research Projects Activity (IARPA), via [2022-21102100005]. The views and conclusions contained herein are those of the authors and should not be interpreted as necessarily representing the ofﬁcial policies, either expressed or implied, of ODNI, IARPA, or the U.S. Government. The US. Government is authorized to reproduce and distribute reprints for governmental purposes notwithstanding any copyright annotation therein.
 
{\small
\bibliographystyle{ieee}
\bibliography{egbib}

\begin{thebibliography}{10}\itemsep=-1pt

\bibitem{anantrasirichai2013atmospheric}
Nantheera Anantrasirichai, Alin Achim, Nick~G Kingsbury, and David~R Bull.
\newblock Atmospheric turbulence mitigation using complex wavelet-based fusion.
\newblock {\em IEEE Transactions on Image Processing}, 22(6):2398--2408, 2013.

\bibitem{aubailly2009automated}
Mathieu Aubailly, Mikhail~A Vorontsov, Gary~W Carhart, and Michael~T Valley.
\newblock Automated video enhancement from a stream of
  atmospherically-distorted images: the lucky-region fusion approach.
\newblock In {\em Atmospheric Optics: Models, Measurements, and
  Target-in-the-Loop Propagation III}, volume 7463, page 74630C. International
  Society for Optics and Photonics, 2009.

\bibitem{belen2001turbulence}
Mikhail~S Belen’kii, John~M Stewart, and Patti Gillespie.
\newblock Turbulence-induced edge image waviness: theory and experiment.
\newblock {\em Applied optics}, 40(9):1321--1328, 2001.

\bibitem{chak2018subsampled}
Wai~Ho Chak, Chun~Pong Lau, and Lok~Ming Lui.
\newblock Subsampled turbulence removal network.
\newblock {\em arXiv preprint arXiv:1807.04418}, 2018.

\bibitem{chen2014detecting}
Eli Chen, Oren Haik, and Yitzhak Yitzhaky.
\newblock Detecting and tracking moving objects in long-distance imaging
  through turbulent medium.
\newblock {\em Applied optics}, 53(6):1181--1190, 2014.

\bibitem{choi2021ilvr}
Jooyoung Choi, Sungwon Kim, Yonghyun Jeong, Youngjune Gwon, and Sungroh Yoon.
\newblock Ilvr: Conditioning method for denoising diffusion probabilistic
  models.
\newblock {\em arXiv preprint arXiv:2108.02938}, 2021.

\bibitem{deng2019arcface}
Jiankang Deng, Jia Guo, Niannan Xue, and Stefanos Zafeiriou.
\newblock Arcface: Additive angular margin loss for deep face recognition.
\newblock In {\em Proceedings of the IEEE/CVF Conference on Computer Vision and
  Pattern Recognition}, pages 4690--4699, 2019.

\bibitem{deng2019retinaface}
Jiankang Deng, Jia Guo, Zhou Yuxiang, Jinke Yu, Irene Kotsia, and Stefanos
  Zafeiriou.
\newblock Retinaface: Single-stage dense face localisation in the wild.
\newblock In {\em arxiv}, 2019.

\bibitem{dhariwal2021diffusion}
Prafulla Dhariwal and Alexander Nichol.
\newblock Diffusion models beat gans on image synthesis.
\newblock {\em Advances in Neural Information Processing Systems},
  34:8780--8794, 2021.

\bibitem{furhad2016restoring}
Md~Hasan Furhad, Murat Tahtali, and Andrew Lambert.
\newblock Restoring atmospheric-turbulence-degraded images.
\newblock {\em Applied optics}, 55(19):5082--5090, 2016.

\bibitem{heusel2017gans}
Martin Heusel, Hubert Ramsauer, Thomas Unterthiner, Bernhard Nessler, and Sepp
  Hochreiter.
\newblock Gans trained by a two time-scale update rule converge to a local nash
  equilibrium.
\newblock {\em Advances in neural information processing systems}, 30, 2017.

\bibitem{hirsch2010efficient}
Michael Hirsch, Suvrit Sra, Bernhard Sch{\"o}lkopf, and Stefan Harmeling.
\newblock Efficient filter flow for space-variant multiframe blind
  deconvolution.
\newblock In {\em 2010 IEEE Computer Society Conference on Computer Vision and
  Pattern Recognition}, pages 607--614. IEEE, 2010.

\bibitem{ho2020denoising}
Jonathan Ho, Ajay Jain, and Pieter Abbeel.
\newblock Denoising diffusion probabilistic models.
\newblock {\em Advances in Neural Information Processing Systems},
  33:6840--6851, 2020.

\bibitem{hufnagel1964modulation}
RE Hufnagel and NR Stanley.
\newblock Modulation transfer function associated with image transmission
  through turbulent media.
\newblock {\em JOSA}, 54(1):52--61, 1964.

\bibitem{karras2019style}
Tero Karras, Samuli Laine, and Timo Aila.
\newblock A style-based generator architecture for generative adversarial
  networks.
\newblock In {\em Proceedings of the IEEE/CVF conference on computer vision and
  pattern recognition}, pages 4401--4410, 2019.

\bibitem{lau2021atfacegan}
Chun~Pong Lau, Carlos~D Castillo, and Rama Chellappa.
\newblock Atfacegan: Single face semantic aware image restoration and
  recognition from atmospheric turbulence.
\newblock {\em IEEE Transactions on Biometrics, Behavior, and Identity
  Science}, 2021.

\bibitem{9320575}
C.~P. {Lau}, A. {Kumar}, and R. {Chellappa}.
\newblock Semi-supervised landmark-guided restoration of atmospheric turbulent
  images.
\newblock {\em IEEE Journal of Selected Topics in Signal Processing},
  15(2):204--215, 2021.

\bibitem{lau2021semi}
Chun~Pong Lau, Amit Kumar, and Rama Chellappa.
\newblock Semi-supervised landmark-guided restoration of atmospheric turbulent
  images.
\newblock {\em IEEE Journal of Selected Topics in Signal Processing}, 2021.

\bibitem{lau2019restoration}
Chun~Pong Lau, Yu~Hin Lai, and Lok~Ming Lui.
\newblock Restoration of atmospheric turbulence-distorted images via rpca and
  quasiconformal maps.
\newblock {\em Inverse Problems}, 35(7):074002, 2019.

\bibitem{lau2020atfacegan}
Chun~Pong Lau, Hossein Souri, and Rama Chellappa.
\newblock Atfacegan: Single face image restoration and recognition from
  atmospheric turbulence.
\newblock In {\em 2020 15th IEEE International Conference on Automatic Face and
  Gesture Recognition (FG 2020)}, pages 32--39. IEEE, 2020.

\bibitem{liu2015faceattributes}
Ziwei Liu, Ping Luo, Xiaogang Wang, and Xiaoou Tang.
\newblock Deep learning face attributes in the wild.
\newblock In {\em Proceedings of International Conference on Computer Vision
  (ICCV)}, December 2015.

\bibitem{lou2013video}
Yifei Lou, Sung~Ha Kang, Stefano Soatto, and Andrea~L Bertozzi.
\newblock Video stabilization of atmospheric turbulence distortion.
\newblock {\em Inverse Problems \& Imaging}, 7(3):839, 2013.

\bibitem{mao2021accelerating}
Zhiyuan Mao, Nicholas Chimitt, and Stanley~H Chan.
\newblock Accelerating atmospheric turbulence simulation via learned
  phase-to-space transform.
\newblock In {\em Proceedings of the IEEE/CVF International Conference on
  Computer Vision}, pages 14759--14768, 2021.

\bibitem{mei2021ltt}
Kangfu Mei and Vishal~M Patel.
\newblock Ltt-gan: Looking through turbulence by inverting gans.
\newblock {\em arXiv preprint arXiv:2112.02379}, 2021.

\bibitem{meinhardt2014implementation}
Enric Meinhardt-Llopis and Mario Micheli.
\newblock Implementation of the centroid method for the correction of
  turbulence.
\newblock {\em Image Processing On Line}, 4:187--195, 2014.

\bibitem{miller2019data}
Kevin~J Miller, Bradley Preece, Todd~W Du~Bosq, and Kevin~R Leonard.
\newblock A data-constrained algorithm for the emulation of long-range
  turbulence-degraded video.
\newblock In {\em Infrared Imaging Systems: Design, Analysis, Modeling, and
  Testing XXX}, volume 11001, page 110010J. International Society for Optics
  and Photonics, 2019.

\bibitem{mittal2012making}
Anish Mittal, Rajiv Soundararajan, and Alan~C Bovik.
\newblock Making a “completely blind” image quality analyzer.
\newblock {\em IEEE Signal processing letters}, 20(3):209--212, 2012.

\bibitem{nair2022comparison}
Nithin~Gopalakrishnan Nair, Kangfu Mei, and Vishal~M Patel.
\newblock A comparison of different atmospheric turbulence simulation methods
  for image restoration.
\newblock {\em arXiv preprint arXiv:2204.08974}, 2022.

\bibitem{nair2021confidence}
Nithin~Gopalakrishnan Nair and Vishal~M Patel.
\newblock Confidence guided network for atmospheric turbulence mitigation.
\newblock In {\em 2021 IEEE International Conference on Image Processing
  (ICIP)}, pages 1359--1363. IEEE, 2021.

\bibitem{nichol2021improved}
Alexander~Quinn Nichol and Prafulla Dhariwal.
\newblock Improved denoising diffusion probabilistic models.
\newblock In {\em International Conference on Machine Learning}, pages
  8162--8171. PMLR, 2021.

\bibitem{oreifej2012simultaneous}
Omar Oreifej, Xin Li, and Mubarak Shah.
\newblock Simultaneous video stabilization and moving object detection in
  turbulence.
\newblock {\em IEEE transactions on pattern analysis and machine intelligence},
  35(2):450--462, 2012.

\bibitem{pearson1976atmospheric}
James~E Pearson.
\newblock Atmospheric turbulence compensation using coherent optical adaptive
  techniques.
\newblock {\em Applied optics}, 15(3):622--631, 1976.

\bibitem{roggemann2018imaging}
Michael~C Roggemann and Byron~M Welsh.
\newblock {\em Imaging through turbulence}.
\newblock CRC press, 2018.

\bibitem{saharia2021image}
Chitwan Saharia, Jonathan Ho, William Chan, Tim Salimans, David~J Fleet, and
  Mohammad Norouzi.
\newblock Image super-resolution via iterative refinement.
\newblock {\em arXiv preprint arXiv:2104.07636}, 2021.

\bibitem{schwartzman2017turbulence}
Armin Schwartzman, Marina Alterman, Rotem Zamir, and Yoav~Y Schechner.
\newblock Turbulence-induced 2d correlated image distortion.
\newblock In {\em 2017 IEEE International Conference on Computational
  Photography (ICCP)}, pages 1--13. IEEE, 2017.

\bibitem{sohl2015deep}
Jascha Sohl-Dickstein, Eric Weiss, Niru Maheswaranathan, and Surya Ganguli.
\newblock Deep unsupervised learning using nonequilibrium thermodynamics.
\newblock In {\em International Conference on Machine Learning}, pages
  2256--2265. PMLR, 2015.

\bibitem{tyson2015principles}
Robert~K Tyson.
\newblock {\em Principles of adaptive optics}.
\newblock CRC press, 2015.

\bibitem{vorontsov2001anisoplanatic}
Mikhail~A Vorontsov and Gary~W Carhart.
\newblock Anisoplanatic imaging through turbulent media: image recovery by
  local information fusion from a set of short-exposure images.
\newblock {\em JOSA A}, 18(6):1312--1324, 2001.

\bibitem{wang2021towards}
Xintao Wang, Yu Li, Honglun Zhang, and Ying Shan.
\newblock Towards real-world blind face restoration with generative facial
  prior.
\newblock In {\em Proceedings of the IEEE/CVF Conference on Computer Vision and
  Pattern Recognition}, pages 9168--9178, 2021.

\bibitem{xie2016removing}
Yuan Xie, Wensheng Zhang, Dacheng Tao, Wenrui Hu, Yanyun Qu, and Hanzi Wang.
\newblock Removing turbulence effect via hybrid total variation and
  deformation-guided kernel regression.
\newblock {\em IEEE Transactions on Image Processing}, 25(10):4943--4958, 2016.

\bibitem{yasarla2020learning}
Rajeev Yasarla and Vishal~M Patel.
\newblock Learning to restore a single face image degraded by atmospheric
  turbulence using cnns.
\newblock {\em arXiv preprint arXiv:2007.08404}, 2020.

\bibitem{yasarla2021learning}
Rajeev Yasarla and Vishal~M Patel.
\newblock Learning to restore images degraded by atmospheric turbulence using
  uncertainty.
\newblock In {\em 2021 IEEE International Conference on Image Processing
  (ICIP)}, pages 1694--1698. IEEE, 2021.

\bibitem{zamir2021multi}
Syed~Waqas Zamir, Aditya Arora, Salman Khan, Munawar Hayat, Fahad~Shahbaz Khan,
  Ming-Hsuan Yang, and Ling Shao.
\newblock Multi-stage progressive image restoration.
\newblock In {\em Proceedings of the IEEE/CVF Conference on Computer Vision and
  Pattern Recognition}, pages 14821--14831, 2021.

\bibitem{zenke2017continual}
Friedemann Zenke, Ben Poole, and Surya Ganguli.
\newblock Continual learning through synaptic intelligence.
\newblock In {\em International Conference on Machine Learning}, pages
  3987--3995. PMLR, 2017.

\bibitem{zhang2018perceptual}
Richard Zhang, Phillip Isola, Alexei~A Efros, Eli Shechtman, and Oliver Wang.
\newblock The unreasonable effectiveness of deep features as a perceptual
  metric.
\newblock In {\em CVPR}, 2018.

\bibitem{zhu2012removing}
Xiang Zhu and Peyman Milanfar.
\newblock Removing atmospheric turbulence via space-invariant deconvolution.
\newblock {\em IEEE transactions on pattern analysis and machine intelligence},
  35(1):157--170, 2012.

\end{thebibliography}
}

\end{document}